\documentclass{article}

\usepackage[nonatbib, final]{nips_2017}

\usepackage[utf8]{inputenc} % allow utf-8 input
\usepackage[T1]{fontenc}    % use 8-bit T1 fonts
\usepackage{hyperref}       % hyperlinks
\usepackage{url}            % simple URL typesetting
\usepackage{booktabs}       % professional-quality tables
\usepackage{amsfonts}       % blackboard math symbols
\usepackage{nicefrac}       % compact symbols for 1/2, etc.
\usepackage{microtype}      % microtypography

\usepackage{caption}
\usepackage{subcaption} % subfigures
\usepackage{verbatim} % for comments
\usepackage{amsmath} % binom

\usepackage{graphicx}      % graphics
\title{Image Segmentation to Distinguish Between Overlapping Human Chromosomes}

\author{
  R. Lily Hu\\
  UC Berkeley, Salesforce Research\\
  \texttt{lhu@salesforce.com} \\  
  \And
 	Jeremy Karnowski\\
   Insight Data Science\\
   \texttt{jeremy@insightdatascience.com} \\
  \And
 	Ross Fadely\\
   Insight Data Science\\
   \texttt{ross@insightdatascience.com} \\
   \And
 	Jean-Patrick Pommier\\
   https://dip4fish.blogspot.fr/\\
   \texttt{jeanpatrick.pommier@gmail.com} \\
}
\begin{document}
% \nipsfinalcopy is no longer used

\maketitle

\begin{abstract}
In medicine, visualizing chromosomes is important for medical diagnostics, drug development, and biomedical research. Unfortunately, chromosomes often overlap and it is necessary to identify and distinguish between the overlapping chromosomes.  A segmentation solution that is fast and automated will enable scaling of cost effective medicine and biomedical research.  We apply neural network-based image segmentation to the problem of distinguishing between partially overlapping DNA chromosomes. A convolutional neural network is customized for this problem. The results achieved intersection over union (IOU) scores of 94.7\% for the overlapping region and 88-94\% on the non-overlapping chromosome regions.
\end{abstract}

\section{Introduction}

Neural networks are a powerful approach to segmenting images, including for street scenes and biomedical images of tissue. In medicine, visualizing chromosomes is important for medical diagnostics, drug development, and biomedical research. Unfortunately, chromosomes often overlap and it is necessary to identify and distinguish between the overlapping chromosomes. For example, some diseases are associated with particular chromosomes or the existence of more or fewer than the expected number of chromosomes. Challenges to this problem include that the overlapping objects may be nearly identical and that it is arbitrary which object is considered the first object and which one the second. Furthermore, overlapping chromosomes may look like one larger chromosome, may criss-cross, or one may be almost entirely on top of the other. A segmentation solution that is fast and automated will enable scaling of cost effective medicine and biomedical research. Traditional methods of distinguishing between overlapping chromosomes involved printing and cutting out individual chromosomes by hand, thresholding on histogram values of pixels, geometric analysis of chromosome contours, among others, and required human intervention when partial overlaps occur. 

In this work, we apply neural network-based image segmentation to the problem of distinguishing between partially overlapping human chromosomes.\footnote{Code available: \url{https://github.com/LilyHu/image_segmentation_chromosomes}} A convolutional neural network, based on U-Net, is customized for this problem. The model is designed so that the output segmentation map has the same dimensions as the input image. To reduce computation time and storage, the model is also simplified. This is because the dimensions of the input image, the set of potential objects in the image, and the set of potential chromosome shapes, are all small, which reduces the scope of the problem, the required capacity of the model, and thus the modeling needs. Various hyperparameters of the model are explored and tested.

Section \ref{background} outlines the background, Section \ref{data} describes the data and preprocessing, Section \ref{methods} elaborates on the model, Section \ref{results} summarizes the results, and Section \ref{conclusion} concludes with future work. 

\section{Background}\label{background}

\subsection{Cytogenetics and Molecular Cytogenetics}
Cytogenetics is the study of chromosomes, including their numbers and structures up to the nucleotids scale \cite{pmid25810762} \cite{pmid26807150} . Pionneering works in species from flies to maise \cite{pmid17246573} enabled the understanding of genes and their inheritance. Human cytogenetics started in 1956 with the discovery of the exact number of chromosomes in humans \cite{pmid12360235}, soon followed by the discovery that structural chromosomal or number anomalies can be be associated with cancer or developmental diseases. Human cytogenetics become a diagnostic tool. Cytogenetics is also used as a biological dosimeter in radiobiology, which is the study of the effect of radiation on living beings \cite{pmid18515111}.

\subsection{Digital Image Processing in Cytogenetics}
The advent of molecular cytogenetics and fluorescent probes (FISH or Fluorescent in-situ Hybridization) yields insights otherwise inaccessible by stained-based cytogenetics. Computers and dedicated software applications started to replace scissor cutouts of black and white pictures of chromosomes for karyotyping.  New algorithms and application were developed to process and interpret fluorescent images, study genomic hybridization, and measure the telomere length Q-FISH \cite{pmid8733138} \cite{pmid10404142} \cite{pmid21663926}. Quantitative methods were developed to become metaphase-free and array-based \cite{pmid25810762}. Metaphasic chromosomes were used to detect targeted chromosomal anomalies \cite{pmid12360235} or for QFISH \cite{pmid22683684}.

Computer based chromosome segmentation and classification is still an open problem \cite{pmid26676686}, particularly the resolving of overlapping chromosomes. Up to now, approaches rely on geometric approachs based on contour analysis \cite{pmid7851156}, finding a skeleton \cite{pmid10207655} \cite{pmid23959611} \cite{1112.4164}. These methods can be rule-based or involve classifiers with hand crafted features. Even for a case as simple as a pair of crossing chromosomes forming a cross, there is ambiguity when it comes to reassembling the pieces to reconstitute the two chromosomes \cite{overlapping_chromosomes_four_points}. Grisan \textit{et al.}  developed a tree search to address this issue \cite{pmid19193514}.

\subsection{Contour-Based Resolution of Crossing Chromosomes}
Chromosomes can be DAPI stained in fluorescence imaging, or stained with giemsa in conventional cytogenetics. After adaptive thresholding and labeling of connected components of binary particles, images of chromosomes can be isolated. Those images can yield single chromosomes, touching chromosomes or overlapping chromosomes. 

\begin{figure}[!h]%tp
   \centering
	\begin{subfigure}[t]{.222\textwidth}
  		\includegraphics[height=4cm]{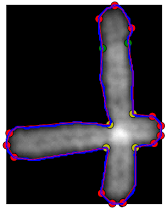}%width=2.1cm,
	\end{subfigure}
	\begin{subfigure}[t]{.2\textwidth}
		\includegraphics[height=4cm]{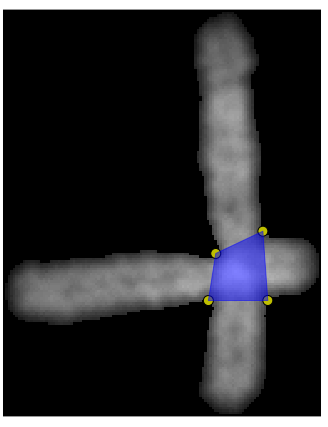}%width=2cm,	
	\end{subfigure}
	\begin{subfigure}[t]{.2\textwidth}
		\includegraphics[height=4cm]{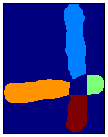}
	\end{subfigure}
	\begin{subfigure}[t]{.2\textwidth}
		\includegraphics[height=4cm]{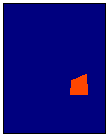}
	\end{subfigure}\\
  
  \caption{Isolation of crossing domain from contour analysis of two crossing chromosomes. Remarkable points are found from contour (left),  crossing domain can be found from four points then used to isolate the different parts of two crossing chromosomes (right).}

\label{contour}
\end{figure}

In the following emblematic example taken from a metaphase, shown in Figure \ref{contour}, a polygonal approximation is computed from the chromosome contour and some remarkable points can be isolated. The four points corresponding to the chromosomal crossing determine a polygon containing the pixels belonging to the overlapping domain.

\begin{comment}
\begin{figure}[!h]%tp
   \centering
	\begin{subfigure}
  		\includegraphics[height=3cm]{Images/contour.png}%width=2.1cm,
	\end{subfigure}
	\begin{subfigure}
		\includegraphics[height=3cm]{Images/FourPoints.png}%width=2cm,	
		\end{subfigure}
	\begin{subfigure}
		\includegraphics[height=3cm]{Images/OverlappDomain-label.png}%width=4.5cm,
  %Images/OverlappDomain-label.png
  
	\end{subfigure}\\
  
  \caption{Isolation of crossing domain from contour analysis of two crossing chromosomes. Remarkable points are found from contour (left),  crossing domain can be found from four points then used to isolate the different parts of two crossing chromosomes (right).}

\label{contour}
\end{figure}
\end{comment}

\begin{comment}
\begin{figure}[!h]%tp
   \centering
  \sbox0{\includegraphics[height=2cm,keepaspectratio]{Images/contour.png}}%width=2.1cm,
  \sbox1{\includegraphics[height=2cm,keepaspectratio]{Images/FourPoints.png}}%width=2cm,
  \sbox2{\includegraphics[height=2cm,keepaspectratio]{Images/OverlappDomain-label.png}}%width=4.5cm,
  %Images/OverlappDomain-label.png
  \begin{subfigure}{\wd0}\usebox0\end{subfigure}%
  \begin{subfigure}{\wd1}\usebox1\end{subfigure}
  \begin{subfigure}{\wd2}\usebox2\end{subfigure}\\
  
  \caption{Isolation of crossing domain from contour analysis of two crossing chromosomes. Remarkable points are found from contour (left),  crossing domain can be found from four points then used to isolate the different parts of two crossing chromosomes (right).}
\end{figure}
\end{comment}

\begin{figure}[!h]%tp
   \centering
   \begin{subfigure}[t]{.9\textwidth}
   		\centering
 		\includegraphics[height=3cm]{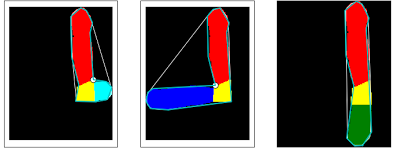}%width=3cm,
 	\end{subfigure}
 	\begin{subfigure}[t]{.9\textwidth}
 		\centering
		\includegraphics[height=3cm]{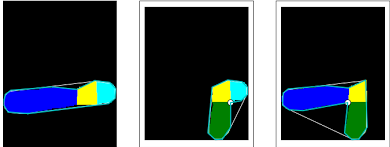}%width=3cm,
	\end{subfigure}
	
  \caption{Combinatorial issue when reassembling segmented parts of two crossing chromosomes. In this case three pairs, mutually exclusive, can be generated. }
  \label{ambiguity}
\end{figure}

Even for a case as emblematic as a pair of crossing chromosomes forming a four-armed cross, there is ambiguity of a combinatorial nature when it comes to reassembling the pieces to reconstitute the two chromosomes \cite{overlapping_chromosomes_four_points}. This ambiguity is illustrated in Figure \ref{ambiguity}.

This ambiguity necessites a decision. Grisan \textit{et al.}  developed a tree search from high resolution Q banded chromosomes to address this issue \cite{pmid19193514}. Successful results were reported on resolving chromosomes clusters\cite{pmid23959611} \cite{1112.4164}, on limited numbers of chromosome clusters extracted from images of metaphases, and in some cases on synthetic images combining chromosomes using Adobe CS\cite{pmid23959611}.

\subsection{Deep Learning for Image Segmentation}

Convolutional neural networks are popular for image segmentation. These include fully convolutional network \cite{long2015fully}, dilated convolutions \cite{yu2015multi}, and encoder-decoder architectures \cite{ronneberger2015u} \cite{badrinarayanan2015segnet}. We propose to solve the overlapping chromosome problem by replacing geometric algorithms with methods from deep learning.

\section{Data}\label{data}

%\subsection{Labeled based resolution of two overlapping chromosomes by semantic segmentation: building a dataset is needed}

%We propose to solve the overlapping chromosome problem in the case of two chromosomes by overcoming the need for geometric algorithms.

\subsection{Collection and Generation}

To create a segmentation solution to resolve overlapping chromosomes, we built a dataset for semantic segmentation using thousands of semi-synthetically generated overlapping chromosomes.

\begin{figure}[hbtp]
\centering
\includegraphics[scale=1]{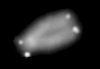}
\caption{Combination of DAPI (Chromosome) and Cy3 (Telomeres) images into a grey scaled image}
\label{greyscale}
\end{figure}

Images of single chromosomes were extracted from an image of human metaphase hybridized with a Cy3 fluorescent telomeric probe \cite{pmid8733138}. Blue (DAPI) and orange (Cy3) components of the image of a single chromosome were combined into a greyscale image as shown in Figure \ref{greyscale}. Then the resolution of the images were decrease by two.

From the set of 46 chromosomes, there are $ \binom{46}{2}=1035$ possible pairs of chromosomes. 12 chromosomes were kept to generate a subset of $ \binom{12}{2}=66$ pairs of chromosomes to combine different chromosomal size and morphology. In each pair of chromosomes, each chromosome was rotated and one chromosome was relatively translated horizontally and vertically to the other one. The overlapping chromosomes were generated by meaning the two grey scaled images of the chromosomes. The so-called ground-truth labels were generated by adding the mask of each single chromosome. By choosing the value 1 for the mask of the first chromosome and the value 2 for the mask of the other chromosome, the label of the overlapping domain has the value 3. Only pairs with ground-truth containing overlapping domains were kept. Raw images of metaphasic chromosomes, dataset and a jupyter notebook are available from kaggle or from dip4fish blog \cite{SyntheticData}, \cite{kaggle}, \cite{github}.

\begin{comment}
\begin{figure}[!h]%tp
   \centering
	\begin{subfigure}
		\includegraphics[scale=0.4,keepaspectratio]{Images/subsetOverlappingChrom-greyscaled.png}
	\end{subfigure}%
	\begin{subfigure}
  		\includegraphics[scale=0.4,keepaspectratio]{Images/subsetOverlappingChrom-groundtruth.png}
	\end{subfigure}
  \caption{Sample of overlapping chromosomes. This dataset contains 13484 pairs of grey-scaled image(left) and ground-truth labels (right) and is available both from kaggle \cite{kaggle}  or github \cite{github}. }
 
\end{figure}
\end{comment}

\subsection{Description of the Dataset}

\begin{figure}[h]
  \centering
  \includegraphics[width=1\textwidth]{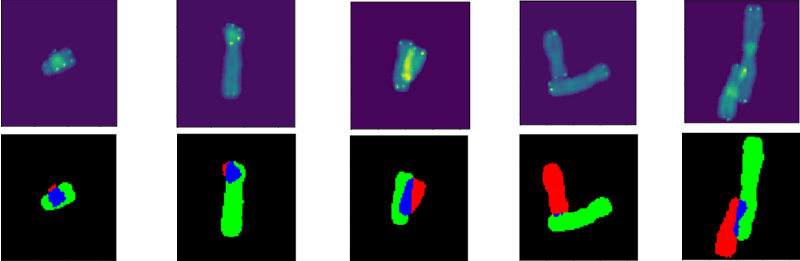}
  \caption{Sample of overlapping chromosomes input image and ground-truth label}
  \label{input}
\end{figure}

The final data set is comprised of about thirteen thousand grayscale images (94 x 93 pixels). For each image, there is a ground truth segmentation map of the same size, as shown in Figure \ref{input}. In the segmentation map, class labels of 0 (shown as black) correspond to the background, class labels of 1 (shown as red below) correspond to non-overlapping regions of one chromosome, class labels of 2 (show as green) correspond to non-overlapping regions of the second chromosome, and labels of 3 (shown as blue) correspond to overlapping regions.

\subsection{Preprocessing}

\begin{figure}[h]
 \centering
	\includegraphics[width=.4\textwidth]{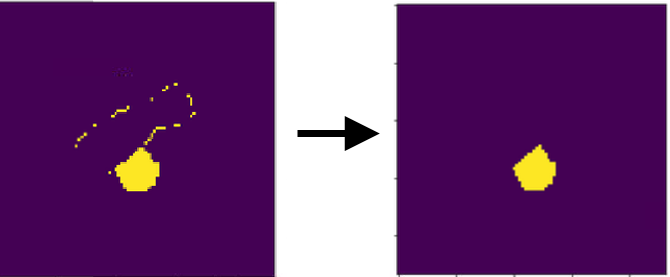}
	\caption{An initial data pre-processing step was performed on segmentation maps that had artifacts}
	\label{data_cleaning}
\end{figure}

A few erroneous labels of 4 were corrected to match the label of the surrounding pixels. Mislabels on the non-overlapping regions, which were seen as artifacts in the segmentation map (example in Figure \ref{data_cleaning}), were addressed by assigning them to the background class unless there were at least three neighboring pixels that were in the chromosome class. The images were cropped to 88 x 88 pixels so that the dimensions were divisible by 2, which helped processing in the pooling layers of the neural network.

\section{Methods and Model Architecture}\label{methods}

\begin{figure}[h]
  \centering
  \includegraphics[width=.6\textwidth]{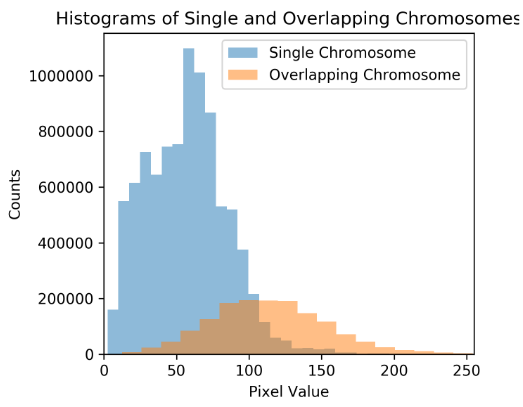}
  \caption{Histogram of pixel vales}
  \label{histogram}
\end{figure}

One simple solution is to classify pixels based on their intensity. Unfortunately, when histograms of the overlapping region and the single chromosome regions are plotted, as shown in Figure \ref{histogram}, there is significant overlap between the two histograms. Thus, a simple algorithm based on a threshold pixel intensity value would perform poorly.

\begin{figure}[h]
  \centering
  \includegraphics[width=1.1\textwidth]{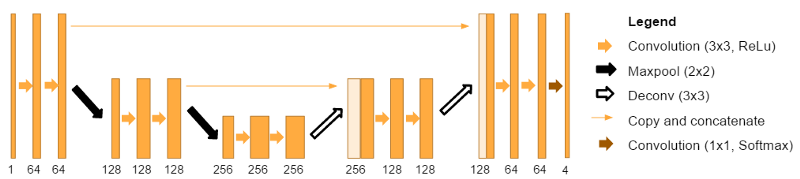}
  \caption{Resulting neural network for separating overlapping chromosomes}
  \label{proposed_nn}
\end{figure}

A convolutional neural network was created for this problem, illustrated in Figure \ref{proposed_nn}. The deep learning solution used for this problem was inspired by U-Net, a convolutional neural network for image segmentation that was demonstrated on medical images of cells. The model for overlapping chromosomes was designed so that the output segmentation map has the same length and width as the input image. To reduce computation time and storage, the model was also simplified, with almost a third fewer layers and blocks. This is because the dimensions of the input image are small (an order of magnitude smaller than the input to U-Net) and thus too many pooling layers is undesirable. Furthermore, the set of potential objects in the chromosome images is small and the set of potential chromosome shapes is also quite limited, which reduces the scope of the problem and thus the modeling needs. Also, cropping was not done within the network and padding was set to be ‘same’. This was because given the small input image, it was undesirable to remove pixels.

Since the problem was not straightforward, various architectures were investigated and the design of the model went through several iterations. These investigations included encoding the class labels as integers, using one-hot encodings, combining the classes of the non-overlapping regions, treating each chromosome separately, using or not using class weights, trying different activation functions, and choosing different loss functions. The model was trained on 64\% of the data, validated on 16\% of the data, and tested on the last 20\% of the data.

\section{Results}\label{results}

\begin{figure}[h]
  \centering
  \includegraphics[width=1\textwidth]{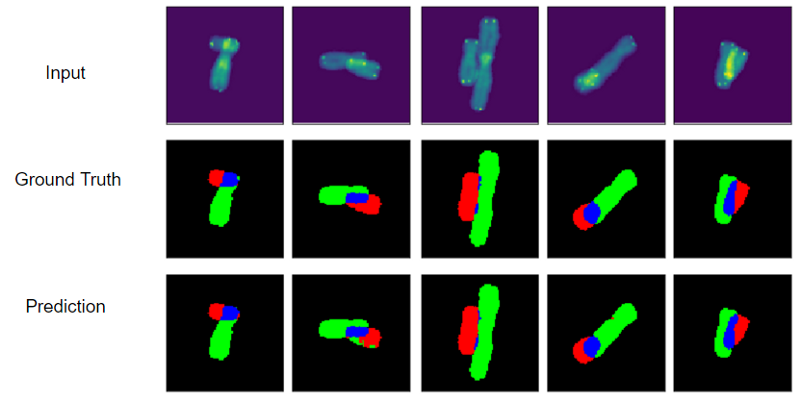}
  \caption{Comparison of prediction with ground truth}
  \label{predictions}
\end{figure}

Visualizations of the input, ground truth, and model predictions are shown in Figure \ref{predictions}. To quantitatively assess the results, the intersection over union (IOU, or Jaccard’s index) is calculated. The model is able to achieve an IOU of 94.7\% for the overlapping region, and 88.2\% and 94.4\% on the two chromosomes. %Graphs of IOU and loss (cross-entropy) versus epoch are plotted in Figure \ref{results_curves}, along with sample predictions in Figure \ref{predictions}. Given that the testing loss is plateauing and not yet increasing, overfitting is not a worry at this training time.

\begin{comment}
\begin{figure}[h]
  \centering
  \includegraphics[width=1.05\textwidth]{images/results_curves.png}
  \caption{Quantitative metrics of model performance}
  \label{results_curves}
\end{figure}

\end{comment}

\section{Conclusion and Future Work} \label{conclusion}

The deep learning model resulted in IOU scores of up to 94.7\% on overlapping chromosomes. To improve the prediction results, the data set can be supplemented with images of single chromosomes and more than two overlapping chromosomes. Data augmentation can also include transformations such as rotations, reflections, and stretching. Additional hyperparameters can also be explored, such as sample weights, filter numbers, and layer numbers. Increasing convolution size may improve misclassification between the red and green chromosomes. %For upsampling, instead of cropping layers, the decoder can use pooling indices computed in the max-pooling step of the corresponding encoder, as in Segnet.

To build a production system that can operate on entire microscope images, the model proposed in this paper can be combined with an object detection algorithm. First, the object detection algorithm can draw bounding boxes around chromosomes in an image. Then, an image segmentation algorithm, based on the model presented here, can identify and separate chromosomes.

\bibliography{biblioChromSegmentation}

\end{document}